\documentclass{getwriting}

\usepackage{amsmath}
\usepackage{graphicx,subfigure}

\title{Investigating the performance of Retrieval-Augmented Generation and fine-tuning for the development of AI-driven knowledge-based systems}

\author[2,3,4]{Róbert Lakatos} 
\author[1]{Péter Pollner} 
\author[2]{András Hajdu}
\author[1,4]{Tamás Joó}

\affil[1]{\footnotesize Data-Driven Health Division of National Laboratory for Health Security, Health Services Management Training Centre, Semmelweis University}
\affil[2]{\footnotesize Department of Data Science and Visualization, Faculty of Informatics, University of Debrecen}
\affil[3]{\footnotesize Doctoral School of Informatics, University of Debrecen}
\affil[4]{\footnotesize Neumann Technology Platform, Neumann Nonprofit Ltd.}

\begin{document}
\nolinenumbers

\maketitle
\begin{abstract}
The development of generative large language models (G-LLM) opened up new opportunities for the development of new types of knowledge-based systems similar to ChatGPT, Bing, or Gemini. Fine-tuning (FN) and Retrieval-Augmented Generation (RAG) are the techniques that can be used to implement domain adaptation for the development of G-LLM-based knowledge systems. In our study, using ROUGE, BLEU, METEOR scores, and cosine similarity, we compare and examine the performance of RAG and FN for the GPT-J-6B, OPT-6.7B, LlaMA, LlaMA-2 language models. Based on measurements shown on different datasets, we demonstrate that RAG-based constructions are more efficient than models produced with FN. We point out that connecting RAG and FN is not trivial, because connecting FN models with RAG can cause a decrease in performance. Furthermore, we outline a simple RAG-based architecture which, on average, outperforms the FN models by 16\% in terms of the ROGUE score, 15\% in the case of the BLEU score, and 53\% based on the cosine similarity. This shows the significant advantage of RAG over FN in terms of hallucination, which is not offset by the fact that the average 8\% better METEOR score of FN models indicates greater creativity compared to RAG.
\end{abstract}

\section{Introduction}
\label{sec:introduction}

Transformer-based large language models (LLMs) are a major advance in natural language processing (NLP). Recently, the most popular networks such as BERT \cite{devlin2018bert}, XL \cite{dai2019transformer}, or GPT \cite{radford2018improving}, and their different versions constantly competed with each other on various language tasks. However, currently,  the best results are achieved by generative large language models (G-LLMs). Among G-LLM, the GPT network, developed by OpenAI \cite{openai} and also operating as part of the ChatGPT \cite{chatgpt} service and the Microsoft Bing search \cite{bingsearch} engine is considered a pioneer. But other competitors also appeared, such as PaLM \cite{chowdhery2022palm} developed by Google, or its further developed version, Gemini \cite{gemini}, which is the basis of Google's Bard system. Furthermore, it is also worth mentioning the LLaMA model. The LLaMA \cite{touvron2023llama} is an open-source G-LLM created by Meta. The most advanced versions of these model families demonstrate remarkable language comprehension and generation capabilities. G-LLMs have transformed the field of natural language processing by achieving next-level performance in various tasks, including text generation, translation, and question-answering. These models are trained on massive datasets of text and code, enabling them to capture complex linguistic patterns and generate human-quality text. However, their true potential often emerges when applied to concrete domains. An example of this is the FunSearch \cite{romera2023mathematical} evolution process. The authors of FunSearch show how language models can even be used to solve mathematical problems.

A specialized field of application G-LLMs is AI-driven knowledge systems, which is a novel approach to the development of such systems. A general and at the same time multimodal approach to AI-driven knowledge-based systems also is the services of OpenAI ChatGPT, Microsoft Bing, or Google Bard. Knowledge-based systems using classic database queries are designed to retrieve and store information in a way that is easy for humans to understand. For this, a combination of NLP and information retrieval techniques (special indexing and querying) are used. In contrast, G-LLMs are suitable for creating new texts, codes, scripts, musical works, e-mails, letters, for example. This allows them to create new information that isn't already in the data they were trained on. This makes them great for tasks such as writing creative text formats, translating languages, or answering complex questions. In turn, these capabilities can be extended to create state-of-the-art knowledge-based systems as well.

To be able to expand the capabilities of G-LLMs and make them suitable for building a knowledge-based system there are two dominant strategies: FN and RAG. In the field of domain adaptation, both solutions offer unique advantages and challenges. 

FN involves further training the G-LLMs on domain and task-specific texts. FN is a well-established technique for domain adaptation of G-LLMs. It involves further training the LLM on a dataset of domain-specific text, allowing the model to incorporate domain-specific knowledge. This approach is effective in improving the performance of G-LLMs on a variety of tasks, including text generation, machine translation, or question-answering.

In the case of RAG, we use a semantic search engine to find the relevant information, which can be injected into the context of the G-LLMs to It can help the model the task solving. Because G-LLMs are sensitive to the transferred context due to the internal attention mechanisms resulting from the transformer architecture. This approach has one of its biggest advantages being that it does not require continuous retraining of the G-LLMs. Namely, it is enough to supplement the database used to generate the context to increase the knowledge base of our system.

Important differences between RAG and FN are that in the case of FN, the risk of hallucination may be greater than in the case of RAG. However, fine-tuned models can better adapt to the target task and reach conclusions that may not be available with RAG. Naturally,  we can apply ensemble approaches. In turn, it is far from trivial whether we can use the ensembled capabilities provided by RAG and FN to achieve better performance.

Currently, there is no recommendation or best practice that precisely defines how to build a knowledge-based system using G-LLMs. This deficiency motivated this chapter of my dissertation. In which I present a possible method of building a knowledge-based system. Considering G-LLM can even be used as an AI-driven expert system in the case of a well-defined target area. This chapter of my thesis is structured as follows. In section \ref{sec:dataset}, we present the databases used to find the best parameters, settings, and methods. In \ref{sec:methodology} section, we describe our methodological approach. In section \ref{sec:results}, we provide the measurement results that demonstrate the performance of the different solutions. Finally, in section \ref{sec:conclusion} we draw my conclusions.

\section{Data}
\label{sec:dataset}

We applied two approaches to create the data. On the one hand, we examined how we can create datasets from PDF and Microsoft Word-based scientific publications because our long-term goals include building our G-LLM-based system. On the other hand, besides the own created data, we created another dataset for the measurements. This second dataset, we composed from publicly available data. All this is to make our results easily reproducible and verifiable.

For the scientific-based dataset collected by us, we curated a collection of specialist publications from urban monitoring and corn cultivation with the help of the National Library of the University of Debrecen and the Faculty of Agriculture, Food Science, and Environmental Management. This corpus, comprising 69 pieces of literature on corn cultivation (CORN) and 83 pieces of literature on urban monitoring (UB), provided a rich source of domain-specific terminology and concepts. Every article or book was available to us in PDF or Word format and the University of Debrecen had to have a special license by which we could download the publications.

As an independent and open-access dataset, we utilized the CORD-19 dataset \cite{wang2020cord}, a freely available repository of tens of thousands of scientific articles on COVID-19, SARS-CoV-2, and related coronaviruses. This dataset encompasses thousands of scientific publications. It is in JSON format and represents about 80 GB of text data. 

The data preparation processes for the model's FN and the RAG application are described in more detail in subsection \ref{sec:dprf}.

\section{Methodology}
\label{sec:methodology}

To decide whether RAG or FN is the better approach for creating a G-LLM-based system, we used the following method. We have determined the models \ref{sec:models} suitable for the task. We have selected the appropriate metrics \ref{sec:metrics}. We prepared the data \ref{sec:dprf} according to the needs of RAG and FN. We fine-tuned the models \ref{sec:finetuning}. Eventually, we evaluated \ref{sec:evaulation} their performance based on the metrics.

\subsection{Models}
\label{sec:models}

To select the models, we took into account the following aspects:

\begin{itemize}
    \item The models must be G-LLM.
    \item The models have been documented scientifically.
    \item The models have pre-trained versions.
    \item The models have been well implemented. That means they should be part of the model repository of the HuggingFace \cite{huggingface} and PyTorch \cite{pytorch} development libraries.
    \item A resource provided by NVIDA DGX Systems should be sufficient to train and test the models since we have access to this platform.
\end{itemize}

Based on these criteria, we selected the GPT-J-6B \cite{gpt-j}, OPT-6.7B \cite{zhang2022opt}, LLaMA-7B \cite{touvron2023llama}, and LLaMA2-7B \cite{touvron2023llama} models.

\subsection{Selected metrics}
\label{sec:metrics}

The following metrics were used to determine the performance of the different language models: Bilingual Evaluation Understudy (BLEU) \cite{papineni-etal-2002-bleu}, Recall Oriented Understudy for Gisting Evaluation (ROUGE) \cite{lin-2004-rouge}, Metric for Evaluation for Translation with Explicit Ordering scores (METEOR) given by \cite{banarjee2005}, and cosine similarity defined in \ref{cosine}.

\begin{equation}
\label{cosine}
Cosine(x, y)= {\frac{xy}{\|x\| \|y\|}} = \frac{ \sum_{i=1}^{n}{x_iy_i} }{ \sqrt{\sum_{i=1}^{n}{(x_i)^2}} \sqrt{\sum_{i=1}^{n}{(y_i)^2}} }
\end{equation}

BLEU is used to measure machine translation performance. BLEU measures n-gram accuracy, which means it counts how many n-grams of the generated text are found in the reference translation.

ROUGE is used to measure the performance of machine translation and text summarization tasks and measures recall, which means that it counts how many n-grams of the reference translation are found in the generated text. ROUGE is designed to work around some of BLEU's limitations. Namely, ROUGE places more emphasis on recall than BLEU and better takes into account the meaning of the text.

METEOR is used to measure the performance of machine translation, text summaries, and creative text formats. METEOR measures Recall, Precision, and word order compatibility.

Cosine similarity is also used to measure the similarity of texts. To use it, the text must be converted into sentence or word vectors and then the cosine similarity between the vectors must be calculated. A higher cosine similarity means that the texts are more similar to each other. This approach We applied so way this by dividing the generated and reference text into sentences and then converting the individual sentences into embedded vectors using the MiniLM L6 v2 \cite{wang2020minilm} sentence transformer. After that, we

\begin{equation}
\label{de:calc_cosine}
CS = Cosine(G \times R) = \left\{ (g,r) | g \in G \text{ and  } r \in R \right\}
\end{equation}

where R is the set of sentence vectors of the reference text and G is the set of sentence vectors of the generated text. Finally, we applied

\begin{equation}
\label{de:max_cosine}
\frac{\sum_{v=1}^{n} \max(CS_v)}{n}
\end{equation}

where $v$ is a similarity value from the cosine similarity matrix $CS$. The maximum of all of $v$ returns the cosine similarity value of the vector where the greatest similarities between the generated sentence and the reference sentence are measured. In other words, we calculated the average of the best matches of the generated sentences with the reference sentences.

\vspace{\baselineskip}

To summarize and compare these metrics:

\begin{itemize}
    \item BLEU is usually the best metric used for machine translation and takes into account matches of words and sentence structures. 
    \item ROUGE is generally the best metric to use for text summaries and creative text formats. 
    \item METEOR is a good general-purpose metric. The meaning and style of the creative text formats such as poems and stories are often evaluated with METEOR. 
    \item With the cosine similarity, we introduce a metric based on vector similarity. With this, we were able to measure the semantic similarity between the referenced and the generated texts.
\end{itemize}

\subsection{Data preparation for RAG and FN}
\label{sec:dprf}

In the case of RAG and FN, we had to use two different approaches to data preparation. In the case of FN, we considered the method of the Stanford Alpaca \cite{alpaca} model to be the guiding principle. In the case of RAG, we have created easily searchable datasets capable of supporting context transfer.

\subsubsection{Q\&A datasets for FN}

To prepare Q\&A datasets, in the matter of collected CORN and UB documents, we split the datasets into paragraphs. In the next step,  we converted them to raw text, and then we cleaned them with the help of human experts. Regarding COVID data, since the entire COVID dataset was too large for our computational resources (NVIDA DGX Systems), so we extracted a subset from it. For this, from the COVID dataset, we selected the articles based on the following filter criteria:

\begin{itemize}
    \item Articles must have abstracts.
    \item Articles must be in the PubMed Central repository. That is, the articles must be open access and medical biology and life science articles.
    \item Articles must have an arxiv id. It also strengthens open access.
    \item Articles must not contain latex elements, so they can also be readable easily and validated by human experts.
\end{itemize}

With these conditions, we managed to restrict the dataset in such a way that we were able to manage it in our environment. We also divided our datasets (CORN, UB) and the COVID dataset into paragraphs. To do this, we took into account the tokenizers of each model. When dividing the paragraphs, we worked in such a way that the individual text parts cannot be longer than 256 tokens according to any model's tokenizer.

To create the questions of the Q\&A dataset, we used the BERT-based generator. The question generator \cite{huggingface_voidful} used by us is available as part of the Huggingface Library's model collection. We generated 5 questions for each paragraph. To be sure that two questions are different for the same paragraph, duplicates were filtered and removed. Thus, we created a maximum of 5 but at least 1 question in the database for each paragraph. With this, we applied a kind of oversampling to the dataset. Table \ref{tab:qaa dataset} lists the number of paragraphs and questions in the created Q\&A ($Q\&A_{CORN}$, $Q\&A_{UB}$, $Q\&A_{COVID}$) datasets:

\begin{table}[ht!]
    \caption{Number of paragraphs and questions of Q\&A datasets}
    \centering
    \vspace{5mm}
    \label{tab:qaa dataset}
    \begin{tabular}{ccc}
        \toprule
        \textbf{Dataset} & \textbf{Paragraphs} & \textbf{Questions} \\ 
        \midrule
        $Q\&A_{CORN}$       & 7058  & 28790 \\ 
        $Q\&A_{UB}$         & 8553  & 27974 \\
        $Q\&A_{COVID}$      & 18004 & 58290 \\
        \bottomrule
    \end{tabular} 
\end{table}

\subsubsection{RAG datasets}
\label{sec:rag dataset}

The performance of the RAG is highly dependent on the accuracy of the context used to answer the questions. Therefore, we used two different approaches to test RAG. On the one hand, we used the $Q\&A_{CORN}$, $Q\&A_{UB}$, $Q\&A_{COVID}$ datasets created for FN. Namely, in this dataset, we managed to generate at least one question for each paragraph. We transformed these questions into vectors using the MiniLM L6 v2 sentence transformer \cite{reimers-2019-sentence-bert}. Thus, with the help of cosine similarity, they became identifiable compared to a reference question. After all, the answers will be in those paragraphs to which the generated questions are most similar to the reference question. It is our first type of indexed dataset ($ID_q$). On the other hand, we also used a more obvious approach.  We split into sentences the entire text and we embedded every sentence with the MiniLM L6 v2 sentence transformer individually. For more effective embedding, sentences shorter than 10 words but longer than 30 words were removed. So we could manage the sentences as vectorized indices. It is our second type of indexed dataset ($ID_s$). In the matter of all datasets (CORN, UB, COVID) we created both types. The properties of these datasets are endorsed in Table \ref{tab:id dataset}.

\begin{table}[ht!]
    \caption{Number of vectorized sentences and questions of indexed datasets}
    \centering
    \vspace{5mm}
    \label{tab:id dataset}
    \begin{tabular}{ccc}
        \toprule
        \textbf{Dataset} & \textbf{Sentences ($ID_s$)} & \textbf{Questions ($ID_q$)} \\ 
        \midrule
        $ID^{CORN}$      & 37874  & 28790  \\
        $ID^{UB}$        & 40002  & 27974  \\
        $ID^{COVID}$     & 56861  & 58290  \\
        \bottomrule
    \end{tabular} 
\end{table} 

\subsubsection{Training, validation and test datasets}
\label{sec:test dataset}

For FN, we split the datasets ($Q\&A_{CORN}$, $Q\&A_{UB}$, $Q\&A_{COVID}$) into training and validation datasets in an 80/20 ratio. We used a special approach for resolution. We did not simply split the datasets, but from those question-answer pairs where we had more than 1 reference question, we selected as many corresponding 20\% of the entire dataset. With this, we achieved that the validation accuracy of the models measures the ability of association of the models. The inference ability of the models was measured on the test dataset. When creating the test dataset, we tried to create questions and related answers that the model definitely could not learn directly. Therefore, we used topic modeling based on nested vectors to create the test dataset. For this, we used Sentence Transformer, UMAP \cite{umap1}, and HDBSCAN \cite{hdbscan} models.

For the identification of the topics, we used $ID_s$ datasets ($ID_s^{CORN}$, $ID_s^{UB}$, $ID_s^{COVID}$).  We embedded with Sentence Transformer all sentences from the $ID_s$ dataset. Following this, we reduced from 386 to 2 the embedded vectors using the UMAP dimension reduction technique. Lastly, we then clustered them with the HDBSCAN algorithm. In the case of HDBSCAN, we set the maximum cluster number to 15 and the minimum cluster number to 6. For the clusters to consist of about 256 tokens for sentences with between 10 and 30 words, the parameters 15 and 6 proved to be optimal. Outlier clusters were removed. We then measured the exact number of tokens contained in each cluster with the tokenizer of each model and, we removed clusters with more than 256 tokens. For the remaining clusters, as in the case of the training and validation datasets, we generated questions here as well. The created test data contained 279 question-answer pairs in each dataset.

\subsection{Fine Tuning settings}
\label{sec:finetuning}

For training, we worked on an NVIDIA DGX computing system. We fine-tuned our models using standard Hugging Face training code with the following belief hyperparameters in the case of all models: loss function to categorical cross-entropy, batch size to 4, learning rate 2e-4, epochs 5, and max length 256.

\subsection{Evaulation strategy}
\label{sec:evaulation}

Our evaluation strategy was to measure ROUGE, BLEU, and METEOR scores for the models. Moreover, we also calculated the cosine similarity of the generated responses compared to the reference responses according to the formulas \ref{de:calc_cosine} \ref{de:max_cosine}. During the evaluation, we followed different strategies for measuring fine-tuned models and RAG.

We fine-tuned the GPT-J-6b, OPT-6.7b, LlaMA-7b, and, LlaMA-2-7b models with the datasets $Q\&A_{CORN}$, $Q\&A_{UB}$, $Q\&A_{COVID}$. For the FN of the models, we measured the validation accuracy at the end of each epoch and saved the models. We only evaluated the best-performing models on the test datasets. To do this, we passed all the questions from the test datasets to the most accurate models. Finally, we calculated BLEU, ROUGE, and METEOR scores and cosine similarity values between the responses generated by the models and the reference responses.

To measure the performance of RAG, we used the LLAMA-2-7b model, which was trained by the authors for this type of application as well. This is not true for several models (GPT-J-6b, OPT-6.7b, LlaMA-7b), so we did not use them to measure RAG performance. In the evaluation of RAG, the context injected and the content of the context are critical. However, the input size of each model may differ. For the model we have chosen, LlaMA-2-7b has a maximum input size of 4096 tokens. The size of the input determines the size of the attachable context. For this reason, we introduced a filter to control the size and quality of the context.

We defined the filtering by a threshold value based on cosine similarity. The threshold value specified what was considered relevant information during the search in the dataset. As described in the section dealing with data preparation, we worked with two types of datasets ($ID_q$, $ID_s$). The measurements were made for all datasets. The threshold values were defined on a scale from 0 to 1 with a step interval of 0.1. This meant that in the case of any question, we discarded matches worse than the threshold value.  Specifically, for example, in the case of a threshold value of 0.5 and a given question taken from the test dataset, only those paragraphs ($ID_q$) or sentences ($ID_s$) passed the filter whose indices showed a cosine similarity greater than 0.5 compared to reference question. This also means that in the case of a threshold of 0, everything, and in the case of a threshold of 1, only the 100\% semantic match is accepted. The sentences ($ID_s$) or paragraphs ($ID_q$) that passed the filter were packaged in a uniform context in descending order of similarity and handed over to the model to try to answer the given question based on it. If the size of the packed context was larger than the input of the given model allowed, the context was cut off at the maximum input size. 

These rules we controlled the size and quality of the context. Using all indexed databases ($ID_q^{CORN, UB, COVID}$, $ID_s^{CORN, UB, COVID}$) we generated answers to all questions in the reference dataset.  Finally, we calculated BLEU, ROUGE, and METEOR scores and cosine similarity values between the responses generated by the models and the reference responses.

\section{Results} 
\label{sec:results}
The FN was carried out as specified in section \ref{sec:finetuning}. During FN, the models achieved the lowest loss on the validation datasets in epochs 3 or 4. 

On the $Q\&A_{CORN}$ and $Q\&A_{COVID}$ the datasets, LlaMA-7b model was the best. On the $Q\&A_{UD}$ dataset, LlaMA-2-7b performed the best. In all cases, the GPT-J-6b and OPT-6.7b models learned with a higher validation loss than the different LlaMA models. The measurements of the FN results are given in Table \ref{tab:id validation} in more detail.

\begin{table}[ht!]
    \caption{The loss of the models measured on the validation datasets by epochs}
    \centering
    \vspace{5mm}
    \label{tab:id validation}
    \begin{tabular}{cllll}
        \toprule
            \multicolumn{1}{r}{\textbf{Epoch}} &
            \multicolumn{1}{r}{\textbf{GPT-J-6B}} &
            \multicolumn{1}{r}{\textbf{OPT-6.7B}} &
            \multicolumn{1}{r}{\textbf{LLaMA-7B}} &
            \multicolumn{1}{r}{\textbf{LLaMA2-7B}} \\
        \midrule
        \multicolumn{5}{c}{$Q\&A_{CORN}$} \\ 
        \hline
        \multicolumn{1}{c}{\textbf{1}} & \multicolumn{1}{l}{0.416608} & \multicolumn{1}{l}{0.596310} & \multicolumn{1}{l}{0.384520} & 0.388159 \\
        \multicolumn{1}{c}{\textbf{2}} & \multicolumn{1}{l}{0.196132} & \multicolumn{1}{l}{0.216316} & \multicolumn{1}{l}{0.162778} & 0.181099 \\
        \multicolumn{1}{c}{\textbf{3}} & \multicolumn{1}{l}{0.163878} & \multicolumn{1}{l}{0.163674} & \multicolumn{1}{l}{0.144640} & 0.149348 \\
        \multicolumn{1}{c}{\textbf{4}} & \multicolumn{1}{l}{0.159600} & \multicolumn{1}{l}{0.153637} & \multicolumn{1}{l}{0.144515} & 0.149162 \\
        \multicolumn{1}{c}{\textbf{5}} & \multicolumn{1}{l}{0.168813} & \multicolumn{1}{l}{0.155910} & \multicolumn{1}{l}{0.154746} & 0.156936 \\ 
        \hline
        \multicolumn{5}{c}{$Q\&A_{UB}$} \\ 
        \hline
        \multicolumn{1}{c}{\textbf{1}} & \multicolumn{1}{l}{0.511894} & \multicolumn{1}{l}{0.825766} & \multicolumn{1}{l}{0.447366} & 3.398389 \\
        \multicolumn{1}{c}{\textbf{2}} & \multicolumn{1}{l}{0.209409} & \multicolumn{1}{l}{0.258804} & \multicolumn{1}{l}{0.180724} & 0.819327 \\
        \multicolumn{1}{c}{\textbf{3}} & \multicolumn{1}{l}{0.170602} & \multicolumn{1}{l}{0.171143} & \multicolumn{1}{l}{0.150210} & 0.186827 \\
        \multicolumn{1}{c}{\textbf{4}} & \multicolumn{1}{l}{0.164166} & \multicolumn{1}{l}{0.159860} & \multicolumn{1}{l}{0.153346} & 0.145882 \\
        \multicolumn{1}{c}{\textbf{5}} & \multicolumn{1}{l}{0.172908} & \multicolumn{1}{l}{0.161635} & \multicolumn{1}{l}{0.162520} & 0.150440 \\ 
        \hline
        \multicolumn{5}{c}{$Q\&A_{COVID}$} \\ 
        \hline
        \multicolumn{1}{c}{\textbf{1}} & \multicolumn{1}{l}{0.586879} & \multicolumn{1}{l|}{1.618626} & \multicolumn{1}{l}{0.678659} & 0.488456 \\
        \multicolumn{1}{c}{\textbf{2}} & \multicolumn{1}{l}{0.238213} & \multicolumn{1}{l|}{0.471962} & \multicolumn{1}{l}{0.218672} & 0.217865 \\
        \multicolumn{1}{c}{\textbf{3}} & \multicolumn{1}{l}{0.192331} & \multicolumn{1}{l|}{0.227678} & \multicolumn{1}{l}{0.182879} & 0.187428 \\
        \multicolumn{1}{c}{\textbf{4}} & \multicolumn{1}{l}{0.186943} & \multicolumn{1}{l|}{0.190710} & \multicolumn{1}{l}{0.185803} & 0.187884 \\
        \multicolumn{1}{c}{\textbf{5}} & \multicolumn{1}{l}{0.194221} & \multicolumn{1}{l|}{0.187811} & \multicolumn{1}{l}{0.195959} & 0.198900 \\
        \bottomrule
    \end{tabular} 
\end{table} 

The final evaluation was performed using the next directives:

\begin{itemize}
    \item We evaluated the models on the test dataset that we presented in subsection \ref{sec:test dataset} of the methodology.
    \item We applied to the measurements the ROUGE, METEOR, BLEU, and CS scores that we presented in section \ref{sec:metrics}.
    \item For the base model LlaMA-2-7b, we also calculated the scores without applying RAG and FN. Since, the creators of the LlaMA-2-7b pre-trained the basic model on a robust corpus, which is a good basis for comparison in the case of FN and RAG. We consider this approach to our evaluation as a baseline.
    \item For each fine-tuned model, we calculated the total score as described in section \ref{sec:evaulation} of the methodology.
    \item In the case of RAG, we calculated the scores using Llama-2-7b base and fine-tuned models as well. 
    \item The threshold value of the search engine used for the RAG presented in section \ref{sec:evaulation} was tested through all possible variations between 0 and 1 with a step interval of 0.1 using the indexed datasets $ID_s$, and $ID_q$.
\end{itemize}

\begin{figure}[ht!]
\centering 
    \subfigure[COVID]{\label{fig:a}\includegraphics[width=120mm]{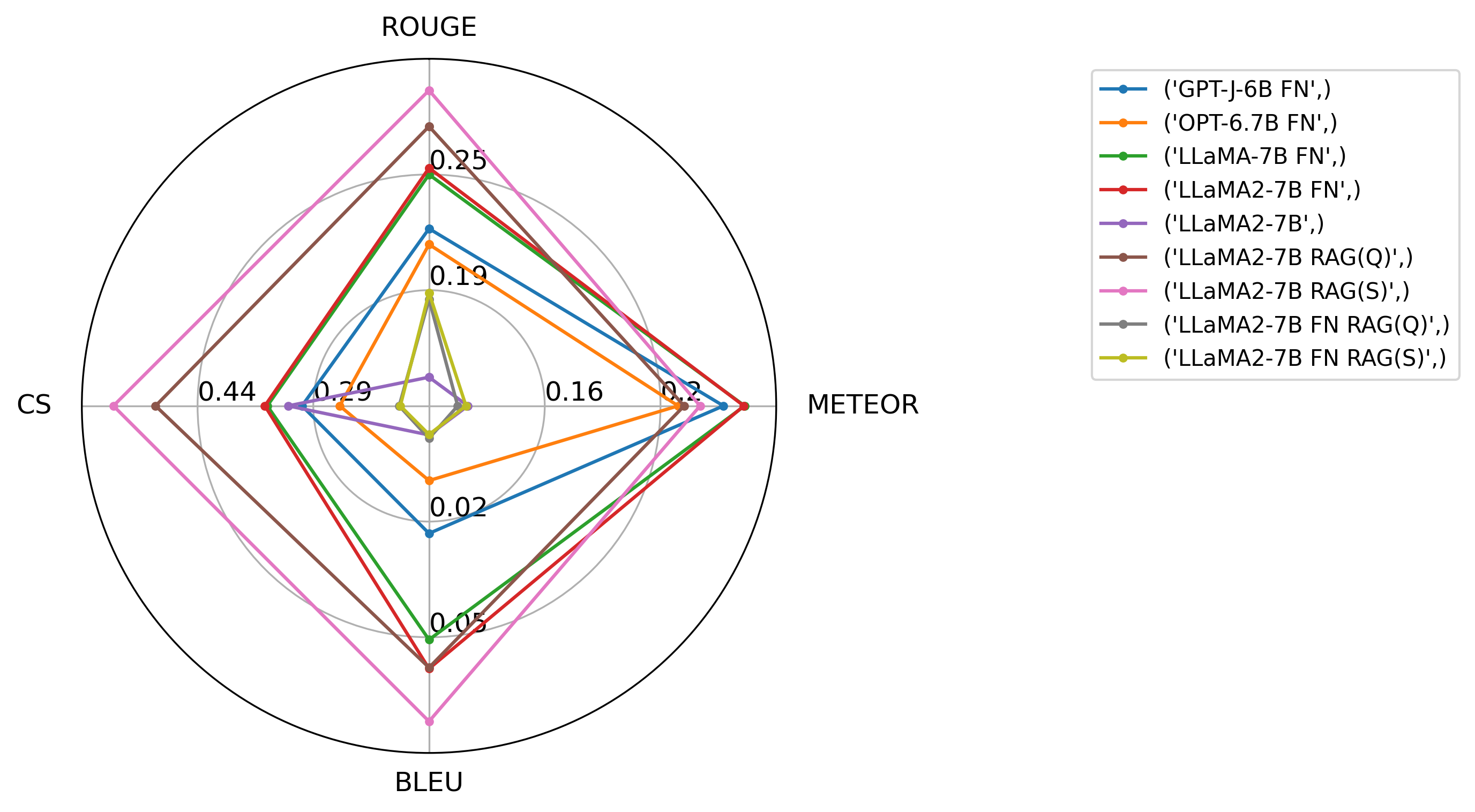}}
    \subfigure[CORN]{\label{fig:b}\includegraphics[width=70mm]{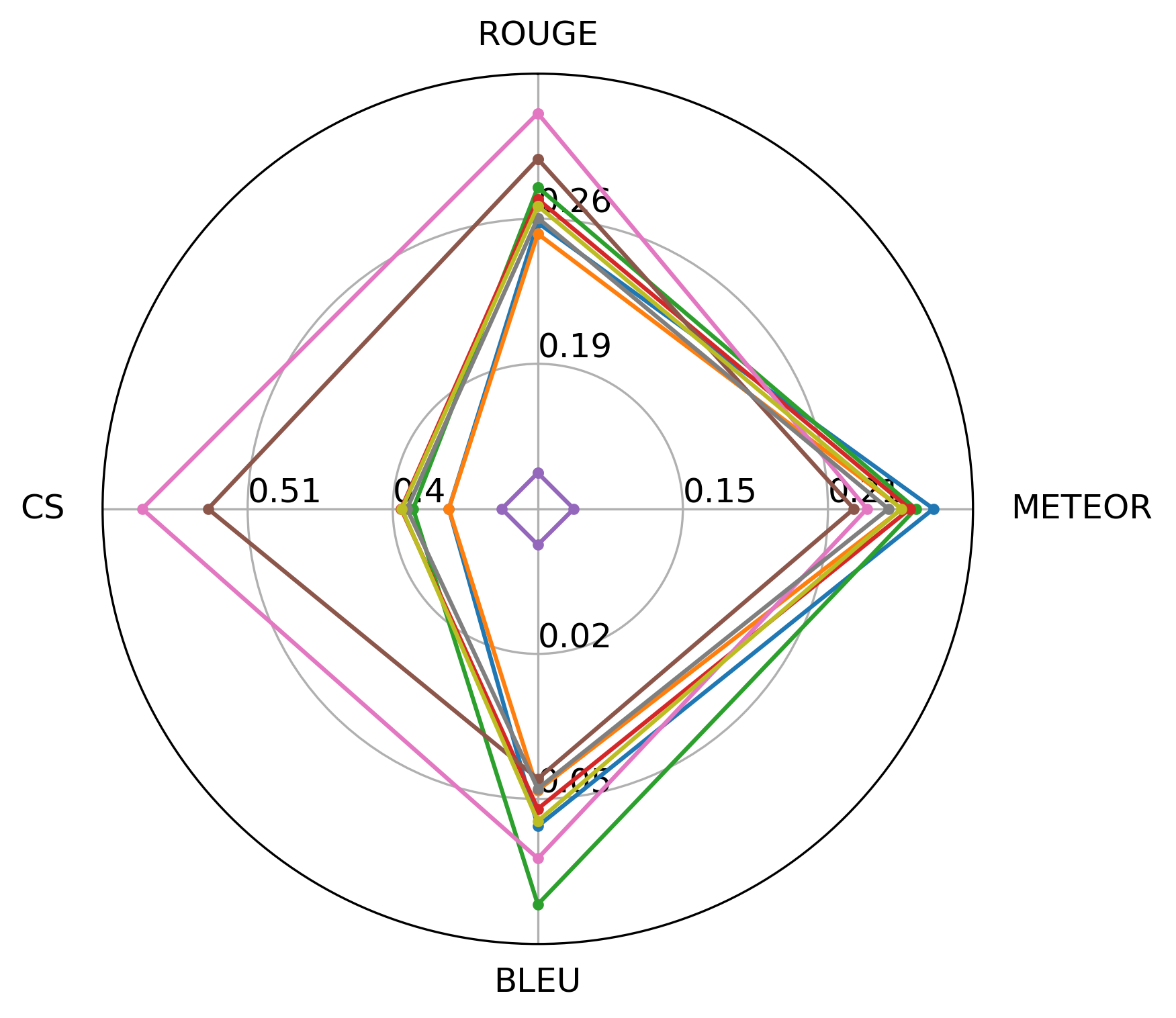}}
    \subfigure[UB]{\label{fig:c}\includegraphics[width=70mm]{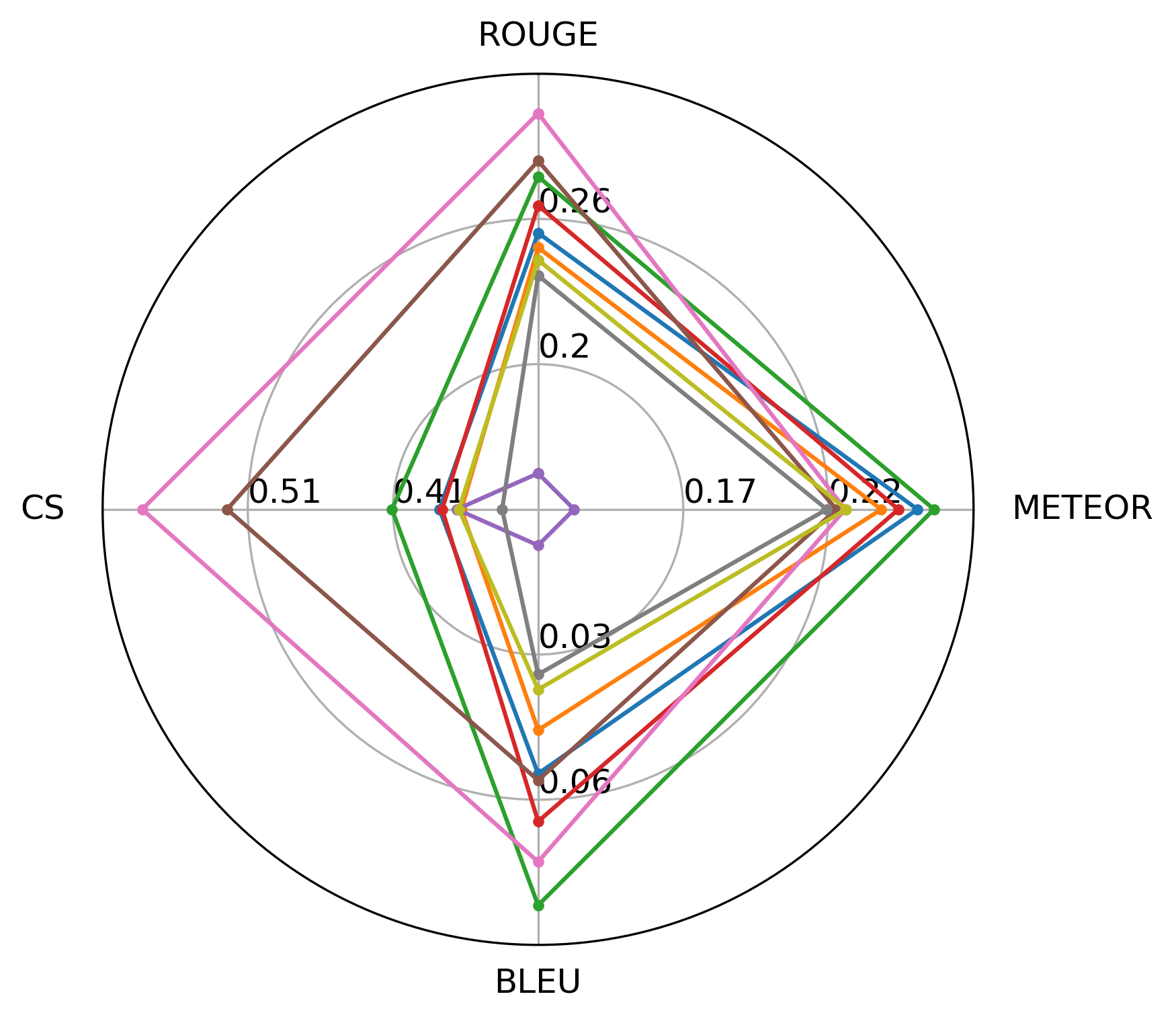}}
    
    \caption{Radar plot of the evaluation results of the models.}
    \label{fig:radar plot}
\end{figure}

We summarize our measurement results in the radar plot in (Figure \ref{fig:radar plot}) which illustrates the relative performance of the models. Furthermore, the average performance of each model approach is presented in Table \ref{tab:avarage results}.

\break

\begin{table}[ht!]
    \caption{Average scores of each approach.}
    \centering
    \vspace{5mm}
    \label{tab:avarage results}
    \begin{tabular}{ccccc}
        \toprule
        \textbf{Models}     & \textbf{ROUGE} & \textbf{METEOR} & \textbf{BLEU} & \textbf{CS} \\ 
        \midrule
        Baseline            & 0.142117           & 0.119251            & 0.002770        & 0.335299        \\
        Fine-tuned          & 0.254003           & 0.242348            & 0.050048        & 0.356439        \\
        RAG with fine-tuned & 0.229296           & 0.195219            & 0.029378        & 0.305797        \\
        RAG                 & 0.294986           & 0.222193            & 0.057998        & 0.544829        \\
        \bottomrule
    \end{tabular} 
\end{table}   	 	 		 	 	 
 	 	 	
As shown in Figure \ref{fig:radar plot} and Table \ref{tab:avarage results}, the results suggest that both FN and RAG outperformed the baseline. RAG performed best and was also the best approach. Moreover, the FN did not help RAG. This is supported by the fact that the best threshold parameter for the LlaMA-2-7b base model during the application of RAG was the value of 0.5. 

In the case of the LlaMA-2-7b finely tuned model, the best threshold was 1.0, which practically means 100\% rejection. So the fine-tuned model could no longer be helped by context injection. The METEOR and BLEU scores of the fine-tuned models were better than those of the RAG models, but in terms of the ROUGE score, they were already inferior compared to the RAG. Furthermore, the RAG produced a significantly better CS score than the fine-tuned models. This shows that RAG significantly improves hallucination and although the association skills of fine-tuned models may be better, the degree of hallucination of fine-tuned models is significantly larger. 

\begin{figure}[!ht]
  \begin{center}          
    \includegraphics[width=300pt]{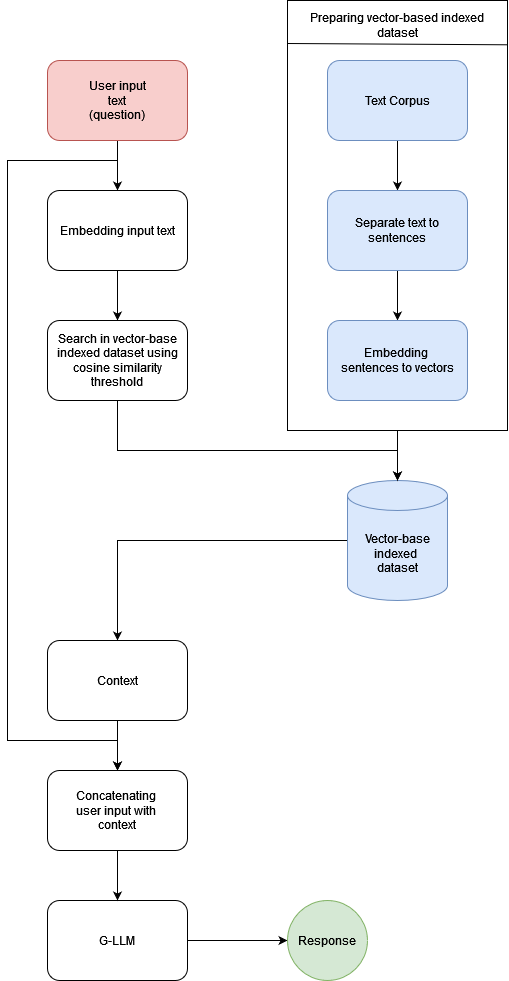}
  \end{center}
  \caption{Flow diagram of the RAG model (best approach) that uses a search engine based on the vectorial embedding of sentences.}
  \label{fig:process}
\end{figure}

Overall, the best result on the test dataset was obtained by using the RAG Llama-2-7b base model with the $ID_s$ dataset. The results of the best approaches are the following: ROUGE 0.3, METEOR 0.22, BLEU  0.063 and, CS 0.57. The best construction is presented in detail in Figure \ref{fig:process}.

\clearpage
\newpage

\section{Conclusions} 
\label{sec:conclusion}

In comparing FN and RAG, RAG achieves better results if we want to create a G-LLM-based knowledge system. In the case of RAG, searching in an indexed database is critical, but by indexing with embedded vectors, it is possible to create a dataset that can be searched efficiently and with which RAG can outperform fine-tuned models. The hallucinations of RAG-based systems are smaller and their expansion is simpler since to expand with new information it is only necessary to add the new data, which requires much less calculation than FN. The combination of FN and RAG is not trivial, as the application of a fine-tuned model with RAG did not result in an extra performance increase.



\bibliography{ref}

\end{document}